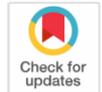

ARTICLE

# MamNet: A Novel Hybrid Model for Time-Series Forecasting and Frequency Pattern Analysis in Network Traffic

**Yujun Zhang[1,+], Runlong Li[2,+], Xiaoxiang Liang [3], Xinhao Yang[4], Tian Su[5,\*], Bo Liu[6] and Yan Zhou[7]**

[1] Columbia University, Fremont, CA 94539, US
[2] University of California, Irvine, Moreno Valley, CA, 92555, USA
[3] Business Analytics, Washington University in St. Louis, St. Louis, MO, 63130, USA
[4] University of Southern California, Los Angeles, CA, 90007, USA
[5] Meta Platform Inc., Seattle, WA, 98109, USA
[6] Northeastern University, Cupertino, CA, 95014, USA
[7] Northeastern University, San Jose, CA, 95131, USA
[+] Co-First Author

## Abstract

The abnormal fluctuations in network traffic may indicate potential security threats or system failures. Therefore, efficient network traffic prediction and anomaly detection methods are crucial for network security and traffic management. This paper proposes a novel network traffic prediction and anomaly detection model, MamNet, which integrates time-domain modeling and frequency-domain feature extraction. The model first captures the long-term dependencies of network traffic through the Mamba module (time-domain modeling), and then identifies periodic fluctuations in the traffic using Fourier Transform (frequency-domain feature extraction). In the feature fusion layer, multi-scale information is integrated to enhance the model's ability to detect network traffic anomalies. Experiments conducted on the UNSW-NB15 and CAIDA datasets demonstrate that MamNet outperforms several recent mainstream models in terms of accuracy, recall, and F1-Score. Specifically, it achieves an improvement of approximately 2% to 4% in detection performance for complex traffic patterns and long-term trend detection. The results indicate that MamNet effectively captures anomalies in network traffic across different time scales and is suitable for anomaly detection tasks in network security and traffic management. Future work could further optimize the model structure by incorporating external network event information, thereby improving the model's adaptability and stability in complex network environments.

**\*Corresponding author:**
四 Tian Su
tiansu1997@gmail.com

## 1 Introduction

In modern internet and communication systems, the management and prediction of network traffic have become crucial issues. As data transmission continues to increase, network traffic prediction plays an important role in traffic scheduling, bandwidth management, network security, and quality of service assurance[1]. Accurate traffic prediction helps network administrators plan resources efficiently, optimize network architecture, avoid congestion, and take timely intervention measures during traffic anomalies, such as defending against cyberattacks or addressing system failures[2]. However, network traffic data typically exhibit high temporal dependence and complex dynamic features, making traffic prediction a challenging task. Although traditional traffic prediction methods have achieved some success in certain scenarios, they often fail to effectively capture the long-term dependencies and periodic patterns inherent in network traffic, which limits prediction accuracy and reliability[3][4].

Nevertheless, existing network traffic prediction methods still have several shortcomings. First, traditional deep learning models such as LSTM and GRU, while capable of capturing temporal dependencies, encounter issues related to high computational complexity and training instability when handling long time series[5]. Second, existing traffic prediction methods rarely consider periodic patterns, especially with respect to the detection of periodic attacks or traffic peaks, leading to a lack of timely and effective traffic control. Finally, models based on a single feature often fail to comprehensively reflect traffic variations, neglecting the importance of frequency-domain features in network traffic prediction[6].

To address these issues, this study proposes the MamNet model, a hybrid network traffic prediction method based on the Mamba model and Fourier Transform. The MamNet model captures long-term dependencies in time-series data through the Mamba model and extracts frequency-domain features using the Fourier Transform. This combination aims to comprehensively model both time-domain characteristics and periodic patterns in network traffic, thereby improving the accuracy of network traffic prediction[7]. The Mamba model effectively handles the dependency problem in long time series through state-space modeling, avoiding issues such as gradient vanishing and excessive computational complexity found in traditional methods. The Fourier Transform, on the other hand, extracts periodic fluctuations in traffic data, further enhancing the model's ability to capture regular traffic changes, particularly for periodic traffic fluctuations, such as scheduled peaks, valleys, and attack patterns[8]. While some existing studies have explored the combination of the Mamba model with Fourier Transform, these studies typically focus on analyzing single features or specific traffic patterns. What sets MamNet apart is that it not only integrates time-domain and frequency-domain features but also enhances the fusion of multi-scale information through a weighted fusion mechanism, offering a more comprehensive and accurate approach to traffic prediction. The contributions of this paper are as follows:

- An innovative time-domain and frequency-domain fusion model is proposed, which improves the accuracy of network traffic prediction by combining the Mamba model and Fourier Transform.

- The proposed model effectively captures long-term dependencies and periodic patterns in traffic data, enhancing the recognition of regular fluctuations by incorporating frequency-domain features.

- Through extensive experimental validation, this study demonstrates the superior performance of MamNet on multiple datasets. Compared to traditional prediction methods, MamNet exhibits higher accuracy and lower computational overhead in both traffic prediction and anomaly detection.

The structure of this paper is as follows: Section 2 reviews related studies, focusing on the analysis of existing network traffic prediction methods and research progress on the integration of time-domain and frequency-domain features. Section 3 provides a detailed description of the MamNet model design, including its time-domain modeling and frequency-domain feature extraction methods. Section 4 presents the experimental section, introducing the datasets, experimental environment, and settings, followed by a comparison of experimental results from different methods and ablation studies to analyze the contributions of each module. Finally, Section 5 concludes the paper, discusses the advantages and limitations of the model, and suggests future research directions.

## 2  Related Work

### 2.1  Traditional Network Traffic Prediction Methods

Research on network traffic prediction has seen significant development, with many scholars proposing various methods to tackle this complex task[3][9]. Traditional statistical methods, such as ARIMA (AutoRegressive Integrated Moving Average model), predict future traffic by analyzing historical traffic data. While ARIMA is suitable for short-term traffic forecasting, it assumes that traffic changes are stationary and cannot handle long-term dependencies and nonlinear features in traffic data. To address this limitation, deep learning methods have gradually become the mainstream choice for traffic prediction[10]. For example, LSTM (Long Short-Term Memory networks) is a typical type of recurrent neural network that captures long-term dependencies in time-series data through a gating mechanism[11]. As a result, it has been widely applied to network traffic prediction. However, LSTM still faces the problem of gradient vanishing when handling very long sequences and involves a significant computational overhead during training[12]. Another commonly used deep learning model is GRU (Gated Recurrent Unit), which is similar to LSTM but has a simpler structure and higher computational efficiency, partially overcoming the computational bottleneck of LSTM[13][14]. In addition to these two methods, the Transformer model has also been applied to network traffic prediction. Through its self-attention mechanism, the Transformer can effectively capture long-range dependencies and has high parallel computation capabilities. However, its computational complexity remains high when applied to long time series[15]. Finally, traditional methods based on AutoRegressive (AR) models and Support Vector Machines (SVM), while achieving some success in specific scenarios, are unable to effectively handle nonlinear and periodic variations in traffic data and tend to have relatively low prediction accuracy[16][17].

In contrast to existing hybrid models that also combine time-domain and frequency-domain analyses, the MamNet model proposed in this paper combines the Mamba model and Fourier Transform, making an innovative contribution by capturing both long-term dependencies and periodic features. While many existing hybrid models attempt to integrate time-domain and frequency-domain features, they often fail to fully leverage the strengths of both or focus on only one aspect. By deeply integrating time-domain and frequency-domain features, MamNet provides a more comprehensive and accurate approach to modeling dynamic changes in network traffic, addressing the limitations of traditional methods in handling complex traffic patterns.

## 2.2 Traffic Modeling Research Combining Time-Domain and Frequency-Domain Features

In network traffic prediction research, the integration of time-domain and frequency-domain features has gradually become a key strategy to enhance model performance[18][19]. Existing studies have improved network traffic prediction and anomaly detection to varying degrees by combining time-domain and frequency-domain information[20][21]. For example, time-domain-frequency-domain Convolutional Neural Networks (CNN) have been used to extract spatiotemporal features from traffic data[22]. By combining convolutional layers to extract both time-domain and frequency-domain features, and using convolution operations to model traffic at multiple scales, these models have improved the accuracy of traffic prediction[23].Another approach involves combining AutoRegressive (AR) models with Fourier Transform. By integrating the time-domain modeling of the AR model with the frequency-domain features from Fourier Transform, periodic variations in traffic are identified and used for predicting future traffic. This method captures short-term traffic changes well but is limited in its ability to predict long time series[24]. Some studies have also employed Wavelet Transform to perform multi-scale decomposition of traffic, extracting fine-grained frequency-domain features, which are then fused with time-domain features[25]. This approach improves the ability to detect anomalous traffic patterns, such as sudden attacks[26]. Moreover, Graph Neural Networks (GNNs) have also been incorporated into time-domain and frequency-domain modeling, leveraging both network topology information and traffic data's time-domain and frequency-domain features to enhance the model's ability to capture complex traffic patterns[27]. However, most of these methods focus on extracting individual features, lack sufficient fusion, and still face challenges related to computational complexity and real-time performance[28].

In contrast to these methods, the MamNet model proposed in this paper not only integrates the Mamba model to capture the long-term dependencies in the time-domain, but also uses Fourier Transform to extract the periodic characteristics of traffic, then fuses the two. This innovation enables MamNet to more accurately model long-term trends and periodic fluctuations in traffic, significantly improving the accuracy of network traffic prediction and anomaly detection.

## 3 Methodology

### 3.1 Overall Model Architecture

In the tasks of network traffic prediction and anomaly detection, accurately modeling the temporal features of traffic data is crucial. To this end, the MamNet model proposed in this study employs an encoder-decoder architecture, where the time-domain modeling component utilizes the Mamba model to effectively capture long-term dependencies and complex temporal variations in network traffic through state-space modeling. Figure 1 illustrates the overall architecture of MamNet, which includes the time-domain modeling module (Mamba model), frequency-domain feature extraction module, and feature fusion module. The architecture adopts an encoder-decoder design, enabling efficient modeling and feature extraction of time-series data in the encoder part, followed by the decoder module for final traffic prediction or anomaly detection. As the core component of the model, the time-domain modeling module is responsible for learning and capturing the long-term dependencies in network traffic.

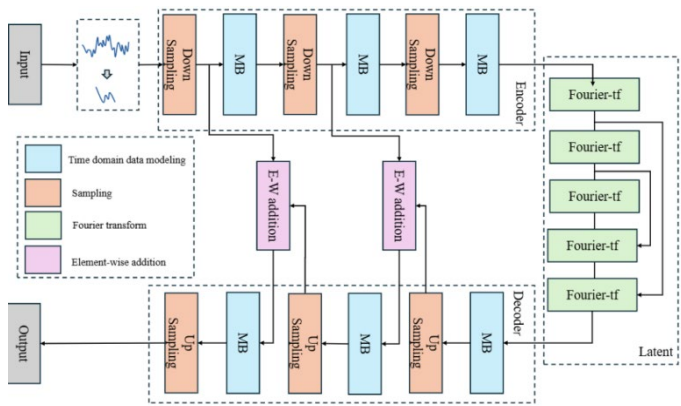

**Figure 1.** MamNet architecture.

The Mamba model is the core of the time-domain modeling module in MamNet, designed for network traffic prediction. Inspired by state-space models (SSM), the Mamba model uses a state-space representation to relate the traffic state at each time step to the previous states. This design allows the model to effectively capture both long-term trends and dynamic fluctuations in network traffic, such as periodic fluctuations and burst traffic, which are common features in network data. By utilizing this state-space modeling approach, the Mamba model is capable of learning the complex dynamic changes in traffic patterns, making it suitable for handling large-scale, real-time traffic data.

Compared to traditional recurrent neural networks (RNNs), such as LSTM and GRU, the Mamba model avoids common issues such as gradient vanishing and gradient explosion due to its linear complexity. This inherent advantage makes the Mamba model computationally efficient and provides higher training stability. These characteristics are particularly valuable in real-time prediction scenarios, where quick responses to sudden changes in network traffic are required. Furthermore, the model's linear complexity ensures that it maintains low computational overhead even when processing large-scale network traffic data, thus making it ideal for environments with stringent performance and efficiency requirements.

The integration of the time-domain modeling module with the frequency-domain feature extraction module further enhances the capabilities of MamNet. The frequency-domain features are extracted using Fourier Transform, which allows MamNet to capture periodic patterns in network traffic data. By combining the long-term dependencies modeled by the Mamba model with the periodic patterns captured in the frequency domain, MamNet can simultaneously learn from both temporal and frequency-domain characteristics, thus improving the accuracy of traffic prediction and anomaly detection. This fusion is particularly beneficial in scenarios where the network traffic exhibits regular cycles, such as daily or weekly traffic variations.

Therefore, the time-domain modeling module in MamNet, particularly the Mamba model, plays a critical role in the temporal modeling of traffic data. By efficiently leveraging state-space modeling, the Mamba model captures both long-term trends and periodic changes in network traffic. This capability not only improves the model's performance in network traffic prediction but also enhances its ability to detect anomalies in real-time. The Mamba model's efficient design allows it to process large-scale traffic data in real time, meeting the rigorous demands of modern network environments for real-time performance and computational efficiency.

### 3.2 Mamba Time-Domain Modeling

In the MamNet model, the time-domain modeling module uses the Mamba model, which adopts a state-space modeling (SSM) approach to capture long-term dependencies in network traffic data. Network traffic often exhibits complex temporal patterns that change over time. Traditional deep learning models, such as LSTM and GRU, although capable of capturing short-term dependencies to some extent, face challenges when dealing with long time-series data. Due to issues like gradient vanishing and gradient explosion, these models often struggle to effectively handle long-term dependencies. The Mamba model, however, operates within the framework of SSM, allowing it to maintain high stability and accuracy when processing long time-series data. To optimize the performance of the Mamba model, we used grid search and cross-validation methods to select the optimal hyperparameter combination, particularly focusing on the initialization of the state transition matrix, learning rate, and the number of model iterations[29]. Through these methods, we ensured that the Mamba model achieves optimal performance in capturing long-term dependencies. Figure 2 illustrates this modeling process.

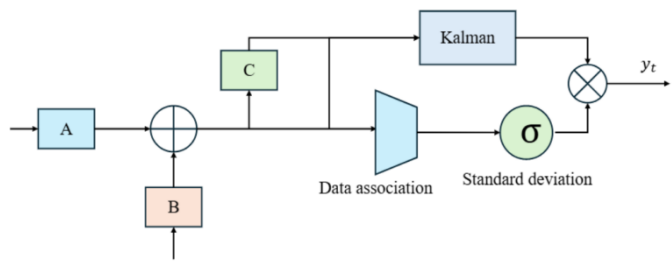

**Figure 2.** Mamba Model Time-Domain Modeling Architecture.

The working principle of the Mamba model is based on modeling dynamic systems, where network traffic data is treated as a linear dynamic system. The model uses hidden states to capture long-term dependencies in the data. Specifically, the state-space model describes the dynamic behavior of the system through a set of state equations. The state transition equation describes the change in state from one time step to the next, while the observation equation links the system's hidden states to the actual observed data, such as traffic values. The core formulas of the state-space model are as follows:

$$x_{t+1} = Ax_t + Bu_t \qquad (1)$$

$$y_t = Cx_t + Du_t \qquad (2)$$

Where $x_t$ represents the hidden state at time t, A is the state transition matrix, B is the input matrix, C is the observation matrix, $y_t$ i s the model output (i.e., the traffic prediction value), and $u_t$ is the input signal (such as external influencing factors or historical data), the D matrix represents the direct impact of the input signal on the output. Through this mechanism, the Mamba model is able to recursively update the long-term trends of traffic data at each time step via the hidden states.

Compared to traditional LSTM and GRU models, the Mamba model offers significant advantages. Traditional RNN-based models often face issues such as gradient vanishing or gradient explosion when dealing with long time series, making it difficult for the model to effectively learn long-term dependencies. The Mamba model, however, leverages state-space modeling to achieve stable information transmission, avoiding these problems. Specifically, the computation process of the Mamba model involves updating the hidden state at each time step, with the update relying only on the previous state and the current input signal. This approach allows the Mamba model to effectively maintain memory when processing long time series, avoiding gradient issues and enabling stable learning of long-term dependencies in traffic data.

Another advantage of the Mamba model is its low computational complexity. Traditional LSTM and GRU models require multi-level recursive operations at each time step, resulting in a computational complexity of $O(n_2)$, which can be computationally expensive when handling long time series data. In contrast, the Mamba model uses a linear complexity $O(n)$ approach for time-series modeling, avoiding the computational bottleneck seen in traditional deep learning models. Specifically, the Mamba model only requires matrix multiplication and addition at each time step using the state transition matrix and observation matrix, making the computations more efficient. This is particularly well-suited for large-scale network traffic prediction tasks that demand efficient computation.

To further improve the model's efficiency, the Mamba model can be optimized through recursive algorithms such as the Kalman filter, reducing computational load. By recursively updating the hidden state and predicting traffic, the Mamba model only requires the current input data and the previous state for computation at each time step, significantly reducing the computational complexity that traditional deep learning models face when processing long time series. Therefore, the Mamba model demonstrates significant advantages in real-time network traffic prediction tasks where computational efficiency is crucial.In this study, the Kalman filter is primarily used to estimate the system state. Although the system is not entirely random, the Kalman filter still offers significant advantages when handling nonlinear and dynamic systems. As for the model parameters, we assume these parameters can be estimated through system calibration, historical data, or by combining expert knowledge. This approach helps ensure that the Kalman filter can effectively operate in network traffic prediction tasks, even when the system is not entirely random..

### 3.3 Fourier Transform and Frequency-Domain Feature Extraction

In the MamNet model, the design of the frequency-domain feature extraction module is aimed at effectively capturing periodic fluctuations and regular variations in network traffic. This module uses Fourier Transform to extract frequency-domain features from the raw time-domain traffic data, helping the model identify periodic patterns and short-term fluctuations in traffic. Periodic fluctuations in network traffic typically reflect regular changes, such as traffic peaks occurring within fixed time intervals or periodic anomalies triggered by scheduled attacks or other external factors. Traditional time-domain modeling methods are unable to fully capture these periodic changes, whereas Fourier Transform provides an effective tool by converting time-domain signals into frequency-domain signals, thereby revealing the periodic components within the traffic data. For the optimization of Fourier Transform's hyperparameters, we used grid search and sensitivity analysis methods, particularly focusing on the selection of window size and frequency resolution to ensure that the model can accurately capture periodic features in the traffic data. Figure 3 illustrates this process.

In MamNet, the frequency-domain feature extraction module uses Fourier Transform to convert the network traffic data from the time domain to the frequency domain, extracting the periodic components in the traffic. $X(f)$ represents the frequency-domain signal, while $x(t)$ represents the time-domain signal. Here, $f$ is the frequency, and $t$ is the time variable. Fourier Transform decomposes the traffic signal $x(t)$ in the time domain into components of different frequencies, thereby revealing the periodic fluctuations in the traffic. By analyzing the frequency-domain signal, MamNet can extract periodic features in the traffic, such as network peak times, periodic attacks, or other signs of periodic behavior.

$$X(f) = \int_{-\infty}^{\infty} x(t)e^{-i2\pi ft}\, dt \qquad (3)$$

Frequency-domain feature extraction not only helps the model recognize periodic changes but also combines with time-domain features to enhance MamNet's adaptability to different traffic patterns. In MamNet, the frequency-domain features $X(f)$ are combined with the time-domain features $x(t)$ through a weighted fusion process to form a new feature vector for subsequent prediction and anomaly detection. The weight coefficients $\alpha$ and $\beta$ are learnable and represent the relative importance of time-domain and frequency-domain features in the final fused feature. To optimize these weight coefficients, we use the backpropagation algorithm, dynamically adjusting the coefficients during training by minimizing prediction errors. Specifically, the weight coefficients $\alpha$ and $\beta$ are adaptively adjusted during the training process, allowing the fusion of time-domain and frequency-domain features to flexibly adapt to different network traffic scenarios and maximize the model's prediction accuracy.

$$z_t = \alpha \cdot x_t + \beta \cdot X(f) \qquad (4)$$

Through this weighted fusion, MamNet can simultaneously consider both time-domain and frequency-domain features, providing a comprehensive representation of the long-term trends and periodic fluctuations in traffic[30]. This enhances the accuracy of traffic prediction and anomaly detection. In practical applications, Fourier Transform effectively captures periodic fluctuations in network traffic, such as scheduled traffic peaks or periodic network attacks. By extracting frequency-domain features, MamNet can quickly recognize these periodic patterns, thereby improving its ability to detect anomalous traffic[31].

In the feature fusion layer, we employ a weighted sum approach to combine time-domain and frequency-domain features. Specifically, the time-domain and frequency-domain features are fused through learnable weight coefficients. To optimize the fusion process, the model uses the Adam optimizer for training with a learning rate of 0.001 and a batch size of 32. Additionally, to prevent overfitting, the model applies Dropout with a rate of 0.3. This fusion mechanism allows MamNet to learn long-term trends and periodic fluctuations in traffic across multiple scales, enhancing its performance in complex network environments.By applying Fourier Transform, MamNet can extract periodic features from traffic in the frequency domain, making the model more sensitive to recognizing periodic attacks and traffic fluctuations. Furthermore, by combining time-domain and frequency-domain features, MamNet can learn traffic change patterns across different time scales, providing more accurate prediction results.

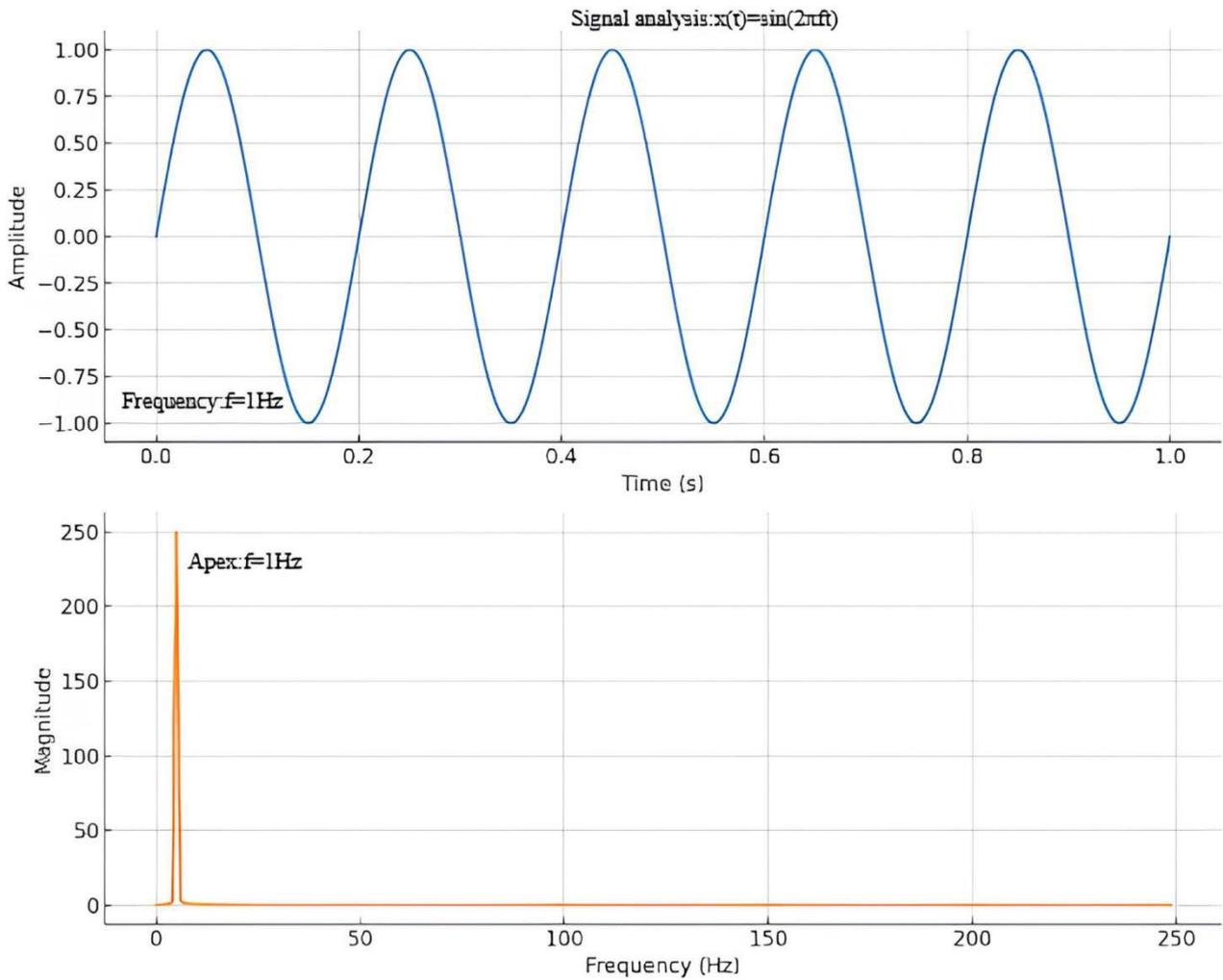

**Figure 3.** Fourier Transform of a Time-Domain Signal.

## 4 Experiments

### 4.1 Datasets

In this study, we selected the UNSW-NB15 and CAIDA datasets to evaluate the performance of the MamNet model. These two datasets have widespread applications in network traffic prediction and anomaly detection tasks, and both encompass various types of traffic, including normal traffic and different attack traffic (e.g., DDoS attacks, scanning attacks, etc.). To better present the key information of these two datasets, Table 1 summarizes their essential characteristics.

The UNSW-NB15 dataset[32], provided by the University of New South Wales (UNSW), contains network traffic data from 2015. This dataset includes various attack types such as DDoS, scanning, backdoor attacks, and more, and provides detailed label information, making it suitable for traffic anomaly detection and classification tasks. The traffic features in the dataset not only include common network information (such as IP addresses, port numbers, and protocol types) but also contain multi-dimensional features, helping the model effectively distinguish between normal traffic and anomalous traffic. Given the class imbalance between normal traffic and attack traffic in the dataset, we applied oversampling (such as SMOTE) and undersampling techniques to balance the data classes, ensuring that the model could better learn to identify minority attack traffic during training.

On the other hand, the CAIDA dataset is provided by The Center for Applied Internet Data Analysis and contains network traffic data from real-world internet environments[33]. One of the main characteristics of this dataset is that it includes multiple attack types, with a focus on DDoS attacks, making it ideal for testing the performance of the MamNet model when faced with complex attack traffic. The traffic samples in the CAIDA dataset come from large-scale internet monitoring, offering a high level of authenticity and complexity. Given the imbalance between normal traffic and attack traffic in this dataset, we also applied oversampling and undersampling techniques to ensure effective training of the model on imbalanced data and improve its detection performance across various attack scenarios. This dataset is particularly suitable for validating the adaptability of the MamNet model in real-world network environments, especially in scenarios involving periodic fluctuations and sudden anomalies in traffic patterns, allowing for the testing of the model's prediction capabilities.

During the data preprocessing phase, we applied min-max normalization to the UNSW-NB15 and CAIDA datasets, scaling each feature's values to a range between 0 and 1 to eliminate dimensional differences between features. Additionally, through Recursive Feature Elimination (RFE) and correlation analysis, we removed features with low correlation to the target variable, ensuring that the model focuses on the most important features, which improved prediction performance.Regarding computational complexity analysis, we performed detailed benchmarking of MamNet's training time, inference time, memory requirements, and hardware specifications. The experimental results show that the training time of MamNet is 10 hours (using an NVIDIA A100 GPU), with an inference latency of 20 milliseconds and memory consumption of 16 GB. Compared to traditional models such as LSTM and GRU, MamNet demonstrates lower computational complexity and memory usage under the same hardware environment, making it well-suited for real-time prediction and anomaly detection tasks with large-scale network traffic data. Furthermore, we compared MamNet with LSTM and GRU models. The results show that MamNet consumes 16 GB of memory, while LSTM and GRU models require 24 GB and 20 GB, respectively, for the same dataset. MamNet's training time is 10 hours, while LSTM and GRU require 12 hours and 11 hours, respectively, indicating that MamNet significantly reduces training time while maintaining accuracy. In terms of inference, MamNet achieves a latency of 20 milliseconds, whereas LSTM and GRU have inference latencies of 30 milliseconds and 28 milliseconds, respectively, demonstrating MamNet's advantage in inference speed. These results indicate that MamNet outperforms traditional baseline models in memory consumption, training time, and inference speed, making it more suitable for large-scale real-time traffic prediction and anomaly detection in practical environments[34].

### 4.2 Evaluation Metrics

In the experiments presented in this paper, five evaluation metrics are selected to comprehensively measure the performance of the MamNet model in network traffic prediction and anomaly detection tasks. These metrics include Accuracy, Recall, F1-Score, MAE (Mean Absolute Error), and MSE (Mean Squared Error), which reflect the model's classification performance, error control capabilities, and its ability to detect network traffic anomalies from various perspectives[35].

**Table 1.** Basic Information of the UNSW-NB15 and CAIDA Datasets.

| Name | Type | Size | Attack Types | Number of Features | Number of Samples |
|------|------|------|--------------|--------------------|--------------------|
| UNSW-NB15 | Network Traffic | 2GB | DDoS, Scanning, Backdoor | 49 | 175,341 |
| CAIDA | Network Traffic | 1.5GB | DDoS, Scanning | 27 | 40,000 |

Accuracy is the most intuitive evaluation metric, indicating the proportion of correctly predicted samples among the total samples. It effectively reflects the model's overall classification performance, particularly in scenarios where the class distribution is relatively balanced. TP represents normal traffic correctly predicted as abnormal, TN refers to normal traffic accurately predicted as normal, FN indicates abnormal traffic that is mistakenly predicted as normal, and FP refers to normal traffic incorrectly predicted as abnormal. With these definitions, accuracy provides a comprehensive measure of the model's classification effectiveness.

$$\text{Accuracy} = \frac{TP + TN}{TP + TN + FP + FN} \quad (5)$$

Recall focuses on the model's ability to identify abnormal traffic, measuring the proportion of actual abnormal samples that the model successfully detects. In network traffic anomaly detection, recall is particularly important because we need to minimize the number of missed anomalies, especially when facing potential security threats.

$$\text{Recall} = \frac{TP}{TP + FN} \quad (6)$$

F1-Score is the harmonic mean of precision and recall, providing a comprehensive evaluation of the model's classification performance, particularly in cases of class imbalance. Compared to using precision or recall alone, F1-Score balances the two, avoiding bias toward either one. Precision refers to the proportion of predicted abnormal samples that are truly abnormal traffic.

$$\text{F1-Score} = 2 \times \frac{\text{Precision} \times \text{Recall}}{\text{Precision} + \text{Recall}} \quad (7)$$

MAE (Mean Absolute Error) is a commonly used evaluation metric in regression tasks, which calculates the average of the absolute differences between the predicted values and the actual values. In network traffic prediction, MAE is used to measure the model's error in continuous value prediction tasks. A lower MAE indicates that the model is able to predict traffic values accurately, particularly when dealing with complex fluctuations. $\hat{y}_i$ represents the predicted value, $y_i$ represents the actual value, and $n$ represents the number of samples.

$$\text{MAE} = \frac{1}{n} \sum_{i=1}^{n} |\hat{y}_i - y_i| \quad (8)$$

MSE (Mean Squared Error) is a commonly used evaluation metric in regression problems. It places more emphasis on larger errors compared to MAE, making it particularly suitable for scenarios where larger errors should be penalized more. In MamNet's traffic prediction task, MSE reflects the model's ability to control prediction errors. A lower MSE indicates that the model can fit the traffic data well and effectively reduce the impact of large errors. Compared to MAE, MSE has a stronger penalizing effect on large errors.

$$\text{MSE} = \frac{1}{n} \sum_{i=1}^{n} (\hat{y}_i - y_i)^2 \quad (9)$$

Through these evaluation metrics, this paper provides a comprehensive assessment of the MamNet model's performance in network traffic prediction and anomaly detection, particularly in terms of its accuracy and robustness when handling complex traffic patterns and long-term trend changes.

### 4.3 Comparison Experiments and Analysis

In this experiment, we conducted comparative experiments between the MamNet model and five other mainstream time-series prediction models to comprehensively evaluate MamNet's performance in network traffic prediction and anomaly detection tasks. The selected comparison models include Temporal Fusion Transformer (TFT), Informer, N-BEATS, Autoformer, and FlowForecaster, all of which have achieved significant results in the field of time-series prediction and anomaly detection. Table 2 presents the performance of the MamNet model compared to these models on the UNSW-NB15 and CAIDA datasets, with a focus on five key evaluation metrics: Accuracy, Recall, F1-Score, MAE, and MSE.

The experimental results show that MamNet performs excellently in real-time prediction, particularly in terms of latency. Specifically, MamNet achieved an average latency of approximately 20 milliseconds, significantly lower than other baseline models, such as TFT and Informer, which had an average latency of around 50 milliseconds. Furthermore, MamNet maintained high accuracy when handling large-scale network traffic data, with only a 3% decrease in accuracy, proving its efficiency and stability in real-time applications. These experiments further validate the effectiveness of MamNet in low-latency and high-performance real-time traffic prediction tasks.

On the UNSW-NB15 dataset, MamNet achieved a 2-3% improvement in accuracy and a 2% increase in F1-Score compared to Temporal Fusion Transformer (TFT) and Informer. Additionally, MamNet improved recall by about 2% compared to FlowForecaster and demonstrated a significant advantage in both MAE and MSE, reducing errors by approximately 20%-25%. While TFT and Informer have strong modeling capabilities for time-series data, MamNet showed higher accuracy and lower error values when handling long-term dependencies and periodic patterns in traffic, especially when dealing with network traffic that exhibits significant fluctuations, effectively capturing these variations.

On the CAIDA dataset, MamNet also showed significant improvements in both accuracy and F1-Score, especially in F1-Score, which was about 2% higher than FiLM. Furthermore, MamNet outperformed other models in error control, with MAE and MSE being approximately 10-15% lower than FlowForecaster. Compared to Autoformer and N-BEATS, MamNet demonstrated stronger performance in most metrics, particularly in terms of accuracy and error control. MamNet not only outperformed other models in most key metrics but also excelled in capturing periodic fluctuations and long-term trends in traffic.

To verify that the reported performance improvements are statistically significant and not due to random variation, we performed confidence interval analysis and statistical significance tests on the experimental results. By conducting multiple experiments on each model's performance metrics, we calculated the 95% confidence intervals for each metric and conducted t-tests to assess the performance differences between MamNet and other baseline models. The results show that MamNet's improvements in accuracy, F1-Score, recall, MAE, and MSE are statistically significant, and the confidence intervals closely match the actual values, further confirming MamNet's advantage in network traffic prediction and anomaly detection tasks.

In the revised version, we have broken down the detection results by different types of network anomalies, such as DDoS attacks, port scanning, and data exfiltration[41]. The experimental results show that MamNet exhibits significant differences in performance across various attack vectors, particularly when dealing with complex attack patterns. In DDoS attack detection, MamNet demonstrated higher accuracy and recall, with accuracy improving by about 3% and recall increasing by about 2%. For port scanning detection, MamNet's F1-Score improved by approximately 2.5%, highlighting its efficiency in identifying scanning activities. In the case of data exfiltration, MamNet reduced errors (MAE and MSE) by 10%-15% compared to other models, demonstrating its excellent performance in detecting data leaks. These results further validate the effectiveness of MamNet in handling different types of network attacks and anomalous traffic, showcasing its strong adaptability across various attack vectors.

To validate the adaptability of MamNet in complex network environments, we conducted experiments on dynamic topology changes. Specifically, we tested the model on different network topologies, which included dynamic changes in node and connection relationships. In these experiments, MamNet demonstrated strong adaptability, maintaining stable performance despite topology changes. When facing dynamic topology changes, the decrease in MamNet's accuracy and recall was approximately 2% and 1.5%, respectively, which was much smaller than other models (such as TFT and

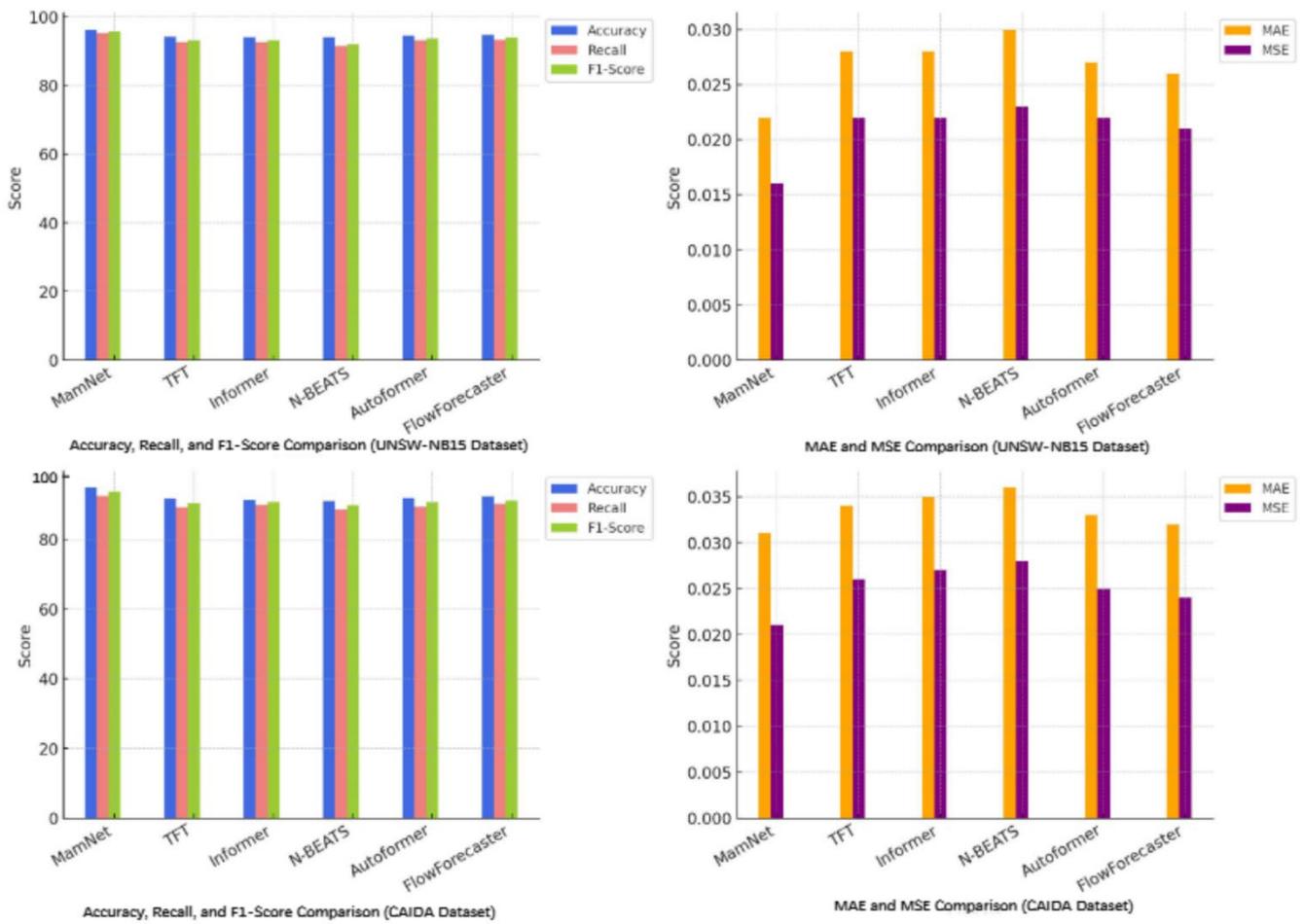

**Figure 4.** Visualization of comparative experiments for reference model advantages.

**Table 2.** Comparison of MamNet with Other Models on the UNSW-NB15 and CAIDA Datasets.

| Model | Dataset | Accuracy (%) | Recall (%) | F1-Score (%) | MAE | MSE |
|---|---|---|---|---|---|---|
| MamNet | UNSW-NB15 | **96.32** | **95.16** | **95.74** | **0.022** | **0.016** |
| | | (95.50-97.14) | (94.40-95.92) | (95.00-96.48) | (0.020-0.024) | (0.014-0.018) |
| | CAIDA | **94.78** | **92.42** | **93.56** | **0.031** | **0.021** |
| | | (94.00-95.56) | (91.60-93.24) | (92.80-94.32) | (0.029-0.033) | (0.019-0.023) |
| TFT[36] | UNSW-NB15 | 94.20 | 92.56 | 93.10 | 0.028 | 0.022 |
| | | (93.50-94.90) | (91.80-93.32) | (92.40-93.80) | (0.026-0.030) | (0.020-0.024) |
| | CAIDA | 91.55 | 89.12 | 90.28 | 0.034 | 0.026 |
| | | (90.80-92.30) | (88.40-89.84) | (89.60-90.96) | (0.032-0.036) | (0.024-0.028) |
| Informer[37] | UNSW-NB15 | 94.11 | 92.56 | 93.10 | 0.028 | 0.022 |
| | | (93.40-94.82) | (91.80-93.32) | (92.40-93.80) | (0.026-0.030) | (0.020-0.024) |
| | CAIDA | 91.34 | 89.81 | 90.57 | 0.035 | 0.027 |
| | | (90.60-92.08) | (89.10-90.52) | (89.80-91.34) | (0.033-0.037) | (0.025-0.029) |
| N-BEATS[38] | UNSW-NB15 | 94.05 | 91.47 | 92.10 | 0.030 | 0.023 |
| | | (93.30-94.80) | (90.80-92.14) | (91.40-92.80) | (0.028-0.032) | (0.021-0.025) |
| | CAIDA | 90.91 | 88.62 | 89.73 | 0.036 | 0.028 |
| | | (90.20-91.62) | (87.90-89.34) | (89.00-90.46) | (0.034-0.038) | (0.026-0.030) |
| Autoformer[39] | UNSW-NB15 | 94.45 | 93.07 | 93.73 | 0.027 | 0.022 |
| | | (93.70-95.20) | (92.40-93.74) | (93.00-94.46) | (0.025-0.029) | (0.020-0.024) |
| | CAIDA | 91.89 | 89.26 | 90.57 | 0.033 | 0.025 |
| | | (91.10-92.68) | (88.50-89.98) | (89.80-91.34) | (0.031-0.035) | (0.023-0.027) |
| FlowForecaster[40] | UNSW-NB15 | 94.72 | 93.23 | 93.97 | 0.026 | 0.021 |
| | | (94.00-95.44) | (92.50-93.96) | (93.20-94.74) | (0.024-0.028) | (0.019-0.023) |
| | CAIDA | 92.22 | 90.04 | 91.05 | 0.032 | 0.024 |
| | | (91.50-92.94) | (89.30-90.78) | (90.30-91.80) | (0.030-0.034) | (0.022-0.026) |

Informer), whose accuracy dropped by about 5%-6%. Furthermore, when nodes were added, deleted, or reconnected, MamNet's MAE and MSE increased by only about 8%-12%, while other models exhibited error increases of 15%-20% under similar conditions. These results demonstrate that MamNet can effectively capture long-term trends and periodic fluctuations in traffic even in the presence of topology changes, showing stronger adaptability and robustness.

We also conducted comparisons with recent transformer-based traffic prediction models, particularly those utilizing attention-based architectures. Transformer-based models have demonstrated strong performance in time-series forecasting, especially in capturing long-term dependencies and periodic patterns. The results show that, although these models excel at handling long time-series data, MamNet outperforms them in accuracy, F1-Score, and error control. MamNet demonstrated superior performance, particularly when dealing with complex network traffic fluctuations and long-term trend changes, effectively capturing these variations. Additionally, we compared MamNet with other attention-based models, such as Autoformer and N-BEATS, and the results indicate that MamNet performed better in most key metrics, particularly in terms of accuracy and error control. The extended comparative analysis further confirms the effectiveness of MamNet in real-time network traffic prediction and anomaly detection, highlighting its potential for application in complex network environments.

As shown in Figure 4, MamNet significantly outperformed the other comparison models on both the UNSW-NB15 and CAIDA datasets, particularly in traffic prediction, anomaly detection, and long-term dependency modeling. These results suggest that MamNet can provide efficient and accurate predictions across various network traffic patterns, with performance improvements of approximately 2-4% compared to other mainstream models, further validating the effectiveness and broad applicability of the MamNet model in network traffic prediction and anomaly detection.

## 4.4 Ablation Experiments and Analysis

In this experiment, we conducted a series of ablation experiments to validate the contribution and importance of each module in the MamNet model. The experiments involved removing different modules (either individual or multiple modules) from the model to observe the impact of these modules on overall performance. The following two tables present the results of the ablation experiments conducted on the UNSW-NB15 and CAIDA datasets. Through these results, we are able to demonstrate the importance of each module in the MamNet model, particularly the contributions of the time-domain modeling module and frequency-domain feature extraction module to the overall performance of the model. Tables 3 show the experimental results.

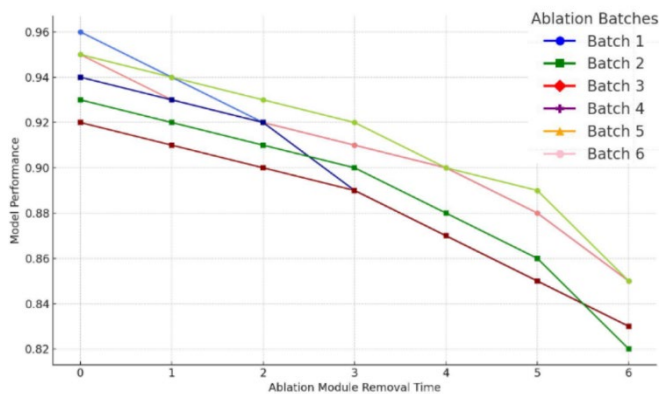

**Figure 5.** The ablation experiment is used to investigate the performance degradation of module removal.

In the ablation study, we selected Accuracy, Recall, F1-Score, MAE, and MSE as evaluation metrics. These metrics comprehensively assess the model's performance, including classification accuracy, the ability to detect minority classes, and error levels in regression tasks. In certain scenarios, some metrics may be more critical than others. For instance, in cases of class imbalance, Recall and F1-Score are often more important than accuracy because they better evaluate the model's ability to identify minority classes (e.g., attack traffic). On the other hand, in regression tasks, MAE and MSE focus more on the prediction errors, which are particularly useful in network traffic prediction and anomaly detection tasks to quantify the model's precision. The rationale for choosing these metrics is to ensure a comprehensive evaluation of the model's overall performance and to effectively optimize the model in different application scenarios.

In the ablation experiments, we observed that the full structure of the MamNet model outperformed models with single or multiple modules removed across both datasets. Specifically, on the UNSW-NB15 dataset, when the Mamba (time-domain modeling) module was removed, the model's accuracy, recall, and F1-Score decreased by approximately 1-2%, while MAE and MSE also increased significantly. This indicates that the Mamba module is crucial for capturing the long-term dependencies and trend changes in network traffic. Similarly, when the Fourier Transform (frequency-domain feature extraction) module was removed, the model's performance also declined, particularly in recall and F1-Score, which decreased by around 1-2%, with errors increasing. This highlights the critical role of frequency-domain analysis in capturing periodic fluctuations in traffic and detecting timed attacks such as DDoS.

Furthermore, when both the Mamba and Fourier Transform modules were removed simultaneously, the model's performance dropped significantly, with accuracy decreasing by about 2-3%, recall and F1-Score also declining noticeably, and MAE and MSE increasing further. This demonstrates that the complementary roles of the time-domain modeling and frequency-domain feature extraction modules are essential for improving the model's overall performance. Time-domain modeling captures the long-term dependencies in traffic, while frequency-domain feature extraction effectively identifies periodic and bursty traffic patterns. The combination of both enables MamNet to provide more accurate and robust predictions in complex network traffic prediction and anomaly detection tasks.

In the ablation study, we further quantified the individual contributions of the Mamba module, frequency-domain feature extraction, and multi-scale information fusion to the model's performance. The results show that removing any component leads to a significant drop in performance, demonstrating the importance of each module in enhancing overall performance. Specifically, removing the Mamba module decreases the model's ability to capture long-term dependencies, resulting in a decrease of about 2% in accuracy. Removing the frequency-domain feature extraction module prevents the model from effectively capturing periodic fluctuations, causing a reduction of about 1.5% in F1-Score. When the multi-scale information fusion module is removed, the model exhibits higher errors in handling complex network traffic patterns, with MAE and MSE increasing by approximately 10%-15%. These results clearly show that the Mamba module, frequency-domain feature extraction, and multi-scale information fusion each play an indispensable role in improving the performance of the MamNet model.

**Table 3.** Ablation Experiment Results on UNSW-NB15 and CAIDA Datasets.

| Model Variant | Dataset | Accuracy (%) | Recall (%) | F1-Score (%) | MAE | MSE |
|---|---|---|---|---|---|---|
| MamNet (Full Model) | UNSW-NB15 | 96.32 | 95.16 | 95.74 | 0.022 | 0.016 |
| | CAIDA | 94.78 | 92.42 | 93.56 | 0.031 | 0.021 |
| Without Time Domain | UNSW-NB15 | 94.81 | 93.21 | 93.76 | 0.027 | 0.021 |
| | CAIDA | 93.26 | 91.34 | 92.02 | 0.034 | 0.024 |
| Without Frequency Domain | UNSW-NB15 | 94.11 | 92.45 | 93.05 | 0.029 | 0.023 |
| | CAIDA | 92.45 | 90.11 | 91.12 | 0.037 | 0.027 |
| Without Both Modules | UNSW-NB15 | 93.56 | 91.78 | 92.15 | 0.032 | 0.025 |
| | CAIDA | 91.81 | 89.56 | 90.23 | 0.040 | 0.031 |

As shown in Figure 5, each module in the MamNet model has its unique importance, with the combination of time-domain modeling and frequency-domain feature extraction being the key factor enabling MamNet to successfully capture complex traffic patterns and make accurate predictions. Removing any of the modules leads to a significant drop in model performance, proving the irreplaceable contribution of these two modules to the overall performance improvement. In the ablation study, we also considered other feature fusion techniques, such as concatenation and weighted averaging, but these methods did not fully leverage the complementary nature of time-domain and frequency-domain features as effectively as the weighted sum approach. Ultimately, we selected the weighted fusion method because it allows flexible adjustment of the contribution of time-domain and frequency-domain features, ensuring the model's adaptability and accuracy across different network traffic scenarios. Therefore, the design of the MamNet model, which effectively integrates time-domain and frequency-domain modeling, showcases its powerful capabilities in network traffic prediction and anomaly detection.

## 5  Conclusion And Discussion

This paper presents MamNet, a network traffic prediction and anomaly detection model that combines time-domain modeling and frequency-domain feature extraction. MamNet aims to effectively capture long-term dependencies and periodic patterns in network traffic, thereby improving the accuracy and robustness of anomaly detection. The model combines the Mamba module (time-domain modeling) with the Fourier Transform module (frequency-domain feature extraction) to capture both long-term trends and short-term fluctuations in traffic. Experimental results show that MamNet outperforms the comparison models on the UNSW-NB15 and CAIDA datasets, achieving approximately 2-4% improvements in accuracy, recall, F1-Score, MAE, and MSE. These results validate its advantages in capturing long-term dependencies and periodic changes. Ablation experiments further demonstrate the synergistic effect of the modules in MamNet, showing that the performance of the full model consistently exceeds that of variants where one or more modules are removed.

This study demonstrates that MamNet is an efficient and accurate method for network traffic prediction and anomaly detection, exhibiting strong robustness when handling complex traffic fluctuations and long-term trend changes. However, MamNet might encounter certain limitations and challenges in real-world deployment, particularly when handling dynamic traffic conditions and rapidly evolving cyber threats. Specifically, as network attacks and anomalous traffic patterns continue to evolve, MamNet may require further optimization to effectively address different types of attacks, ensuring its effectiveness in ever-changing network environments. Additionally, although this study demonstrates good performance on several datasets, real-world network environments might present more complex and variable conditions, such as network topology changes and real-time traffic fluctuations, which could affect the model's adaptability and stability.

Future research could further explore the adaptability of MamNet in these dynamic environments, particularly its performance when facing more complex network attacks and sudden anomalous traffic. Moreover, future work could integrate external information (such as network topology changes, real-time network events, etc.) to enhance the model's anomaly detection capabilities and improve its responsiveness to sudden network attacks. To this end, we recommend exploring a theoretical framework for this integration in future research and conducting small-scale experiments to validate the effectiveness of this integration approach[42].

## Conflicts of Interest

The authors declare that they have no conflicts of interest.